\documentclass{svproc}
\usepackage{url}
\usepackage{orcidlink}
\usepackage{hyperref}

\begin{document}
\mainmatter              
\title{Learning Wall Segmentation
in 3D Vessel Trees using Sparse Annotations}
\titlerunning{Learning Wall Segmentation
in 3D Vessel Trees using Sparse Annotations}  
\author{Hinrich Rahlfs\orcidlink{0009-0005-1870-9290} \inst{1(*)} \and Markus Hüllebrand\orcidlink{0000-0003-4948-0917} \inst{1,2,3} \and 
Sebastian Schmitter\orcidlink{0000-0003-4410-6790} \inst{4} \and Christoph Strecker \inst{5} \and Andreas Harloff\orcidlink{0000-0002-3252-7910} \inst{5} \and Anja Hennemuth\orcidlink{0000-0002-0737-7375} \inst{1,2,3}}
\authorrunning{Hinrich Rahlfs et al.} 
%
\tocauthor{Hinrich Rahlfs, Markus Hüllebrand, 
Sebastian Schmitter\, Christoph Strecker, Andreas Harloff,  Anja Hennemuth}
\institute{Institute of Computer-Assisted Cardiovascular Medicine, Charité - Universitätsmedizin Berlin, Germany\\\email{hinrich.rahlfs@dhzc-charite.de}
\and Fraunhofer MEVIS, Bremen, Germany \and DZHK (German Centre for Cardiovascular Research), Partner Site Berlin, Germany \and Physikalisch-Technische Bundesanstalt, Berlin, Germany \and Department of Neurology and Neurophysiology, Faculty of Medicine, Medical Center—University of Freiburg, University of Freiburg, Germany}

\maketitle              

\begin{abstract}
We propose a novel approach that uses sparse annotations from clinical studies to train a 3D segmentation of the carotid artery wall. We use a centerline annotation to sample perpendicular cross-sections of the carotid artery and use an adversarial 2D network to segment them. These annotations are then transformed into 3D pseudo-labels for training of a 3D convolutional neural network, circumventing the creation of manual 3D masks. 
For pseudo-label creation in the bifurcation area we propose the use of cross-sections perpendicular to the bifurcation axis and show that this enhances segmentation performance. Different sampling distances had a lesser impact. 
The proposed method allows for efficient training of 3D segmentation, offering potential improvements in the assessment of carotid artery stenosis and allowing the extraction of 3D biomarkers such as plaque volume.

\end{abstract}

\section{Introduction}
\begin{figure}
    \centering
    \includegraphics[width=1\linewidth]{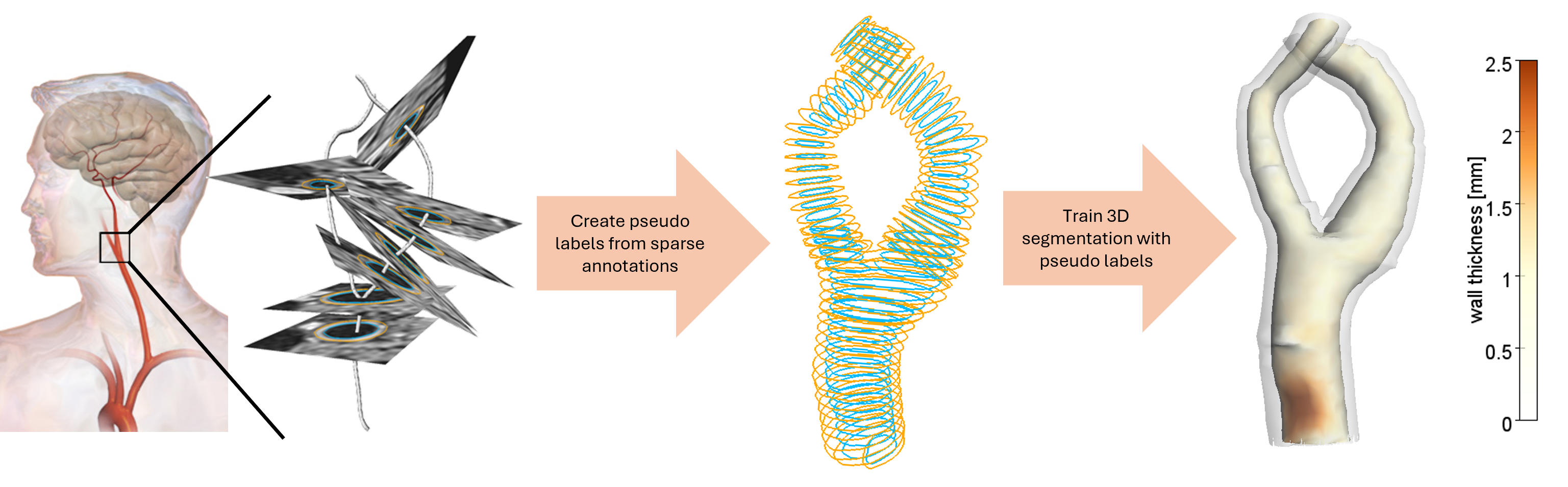}
    \caption{(left) Centerlines were extracted and the vessel wall was annotated in eight 2D cross-sections per carotid artery in 3D BB-MRI images. (adapted from \cite{blausen2014medical}) (middle) The centerline and an adversarial 2D network are used to create densely distributed contours of the carotid artery wall. (right) These contours are used to train a 3D nnU-Net, enabling automatic 3D segmentation of the carotid artery. This allows the extraction of the vessel wall thickness and the visualization of the carotid geometry.}
    \label{fig:training-procedure}
\end{figure}
3D segmentation of the carotid artery wall in black blood (BB)-MRI is of great interest to assess carotid artery stenosis, a major risk factor for stroke \cite{campbell2019ischaemic}. The manual creation of 3D segmentation masks shows high variability between observers and is a tedious task. Annotations are often created slice by slice. \cite{van2012automated}.

Automatic, neural network-based 3D segmentation methods could enable fast reproducible carotid artery analysis, and promising approaches have been published. Zhu et al. \cite{zhu2021cascaded} trained a 3D segmentation of the carotid artery wall with 21 manually annotated BB-MRI scans but achieved only a high mean average surface distance of 0.88~mm for the carotid wall. Lavrova et al. \cite{lavrova2023ur} trained a 3D segmentation with 106 manually annotated BB-MRI scans in which only the symptomatic carotid arteries were manually annotated. Both methods require time-consuming manual creation of 3D masks for network training. Recent publications have introduced concepts for the training of 3D vessel segmentation models with sparse annotations. Hu et al. \cite{Hu2022LabelPropagation} used label propagation from annotations on axial slices to learn a 3D segmentation of the carotid artery. This reduces the annotation time but requires a specific annotation scheme. Namely, annotations need to be equally distributed and on axial slices.
Brosig et al. \cite{brosig2024learning} proposed to create 3D training masks of the aortic root using the annotation of perpendicular cross-sections and the Poisson surface reconstruction.

The goal of our work is to train a neural network directly on clinical annotations, which are typically created on cross-sections perpendicular to the centerline and focus on the carotid bulb. \cite{strecker2020carotid,strecker2021carotid} On the left side of Fig.~\ref{fig:training-procedure}, annotated cross-sections of one carotid artery are shown, and it is clear that label propagation and direct use of the Poisson reconstruction are not possible because the annotated planes are too sparse. For example, only one cross-section in the external carotid artery (ECA) has been annotated.

In a previous study, we showed that these sparse 2D annotations can be used to train a 2D segmentation of the carotid artery wall and that the trained network generalizes well to cross-section position. \cite{rahlfs2024carotid} We propose to use this 2D segmentation to create 3D pseudo-labels for network training. We use the sparse annotations to compare different methods for dealing with the bifurcation area and determine how well the method generalizes to different scan parameters and healthy subjects.

\section{Method}
\begin{table}[h]
\centering
\caption{Patient demographics of test and training set.}
\label{tab:meta_data_training_vs_testing}
\begin{tabular}{l|lll}
                                & Training set  & Test set \\ \hline
$\frac{No. female}{No.male}$    & 0.45          & 0.44\\
$\mu(Age)$ in years             & 70.75         & 70.57\\
$\mu(Weight)$ in $Kg$           & 77.64         & 80.26\\
$\mu(BMI)$ in $\frac{Kg}{m^2}$  & 26.5          & 27.7
\end{tabular}
\end{table}
\paragraph{Data}
For network training and evaluation, we used 202 MRI volumes of 121 subjects with hypertension, at least one additional cardiovascular risk factor, and a plaque in the internal carotid artery (ICA) or common carotid artery (CCA) measuring $\geq$ 1.5~mm in ultrasound. Subjects with ICA-stenosis $>$ 50\% according to NASCET criteria \cite{reutern2012grading} were excluded. The MRI volumes cover both carotid arteries in the region of the carotid bifurcation. The data was acquired on a Siemens Prisma 3T scanner using the 3D Turbo Spin Echo sequence 3D-SPACE with fat saturation and dark-blood preparation. The sequence used $T_R=900ms$, $T_E=26ms$, and an isotropic voxel resolution of 0.6~mm. The data is described in detail by Strecker et al. \cite{strecker2020carotid,strecker2021carotid}.
The training set contains 181 MRI volumes of 108 subjects; the test set contains 21 MRI volumes of 13 subjects. The demographics of the subjects in the training and test sets are similar (see Table~\ref{tab:meta_data_training_vs_testing}). \cite{rahlfs2024carotid}

A stenosis test set was created with 5 test set subjects that showed a stenosis $\ge 10\%$. It contains one cross-section per carotid artery, placed at the maximal wall thickness.

10 additional MRI volumes of young, healthy subjects acquired with the same scanner and MRI sequence were included for model evaluation.

To evaluate generalization to a different annotator, scanner, and sequence, we used the test set of the 2021 Carotid Artery Vessel Wall Segmentation Challenge \cite{Challenge2021}.
\begin{figure}[h]
    \centering
    \includegraphics[width=0.7\linewidth]{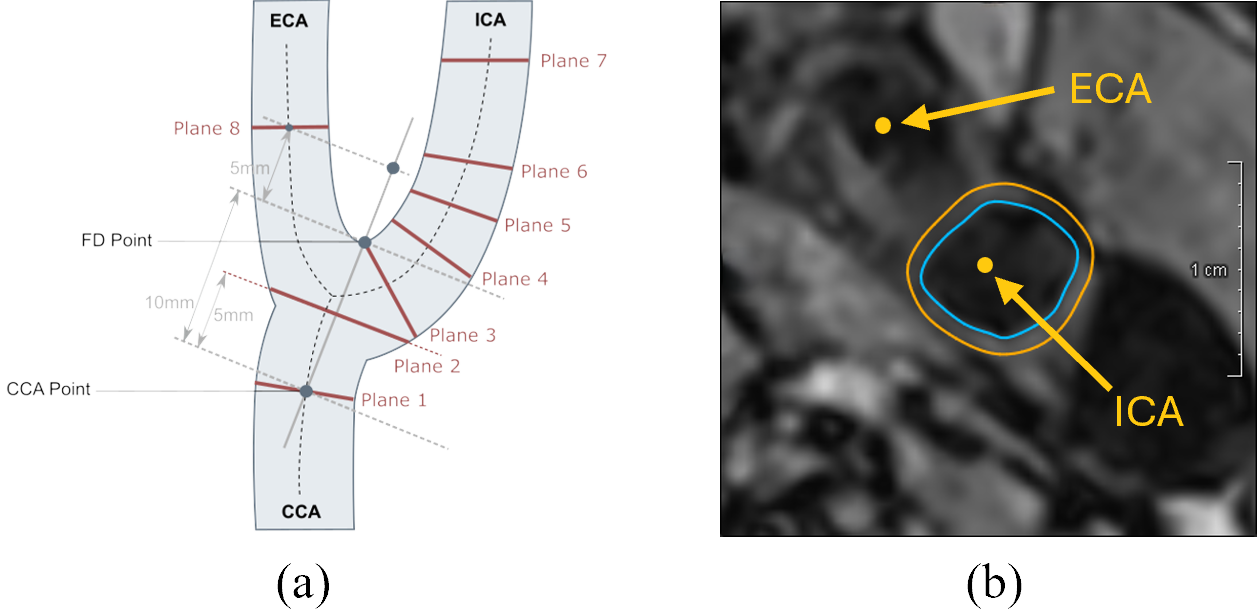}
    \caption{(a) Placement of sparse annotations as proposed by Strecker et al. \cite{strecker2020carotid,strecker2021carotid} (b) Annotation of the inner (cyan) and outer (orange) ICA wall. The ECA is visible in the cross-section but not annotated.}
    \label{fig:sparse-annotations}
\end{figure}
\paragraph{Sparse Annotations}
The sparse annotations used for training and evaluation are the centerline of the carotid artery and eight manually annotated 2D cross-sections per carotid artery (Fig.~\ref{fig:sparse-annotations}). Experts manually annotated the vessel wall of the CCA, ICA or ECA. Fig.~\ref{fig:sparse-annotations}b shows a cross-section with ICA and ECA. Only the vessel wall of the artery in the center was annotated. The process of annotation creation is described in detail by Strecker et al. \cite{strecker2020carotid}

\paragraph{Training of adversarial 2D networks}
In the first step, an adversarial 2D U-Net is trained fully supervised on the 2D cross-sections. The dataset and training procedure are described in detail by Rahlfs et al. \cite{rahlfs2024carotid}

\paragraph{Creation of Pseudo Labels}
We created a 3D pseudo-label mask for all MRI volumes in the training set. To create the pseudo-labels, we sampled 2D cross-sections along the centerline and used the adversarial network to segment them. We then transformed the segmentation into contours (Fig.~\ref{fig:training-procedure}) and the contours into a 3D surface using the Poisson reconstruction. \cite{poisson}

\begin{figure}
    \centering
    \includegraphics[width=1\linewidth]{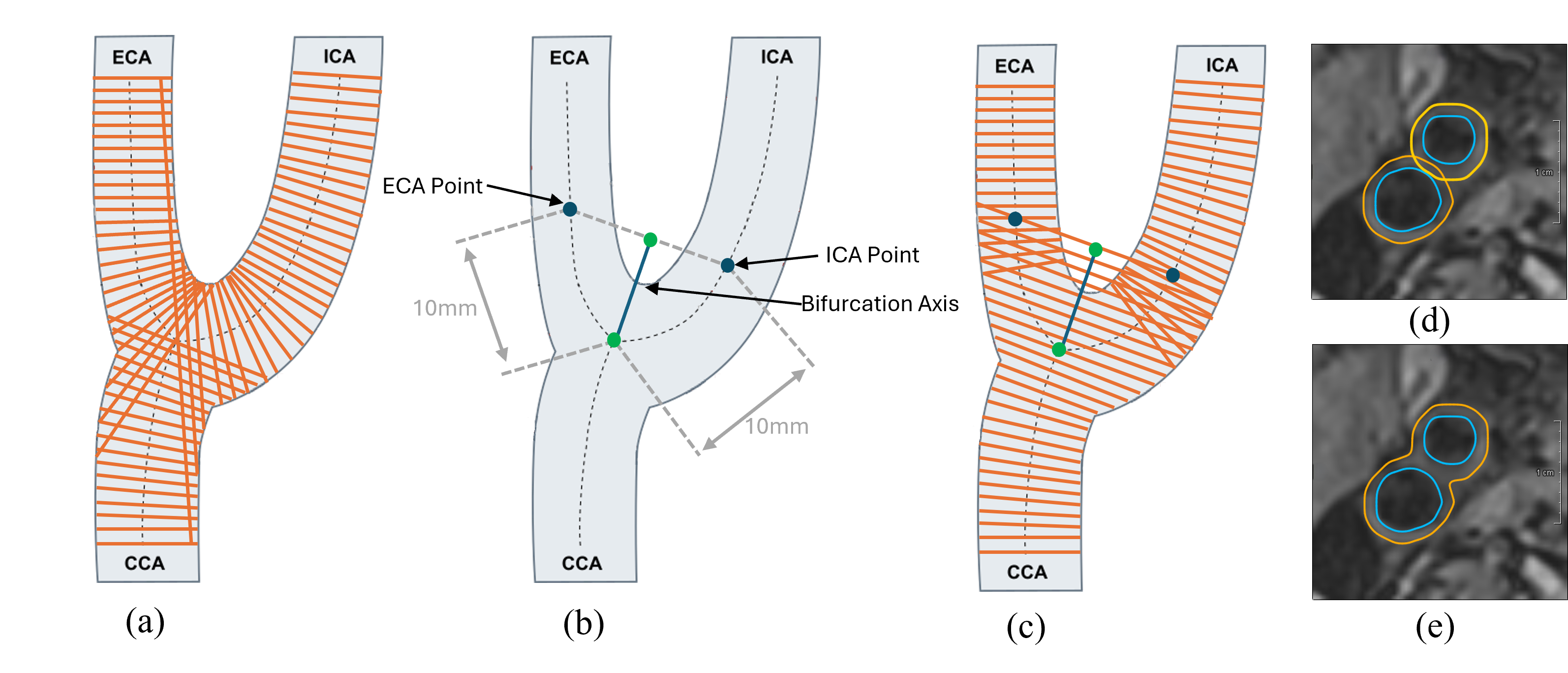}
    \caption{(a) Cross-section sampling perpendicular to the centerline. In the bifurcation area, this can, e.g., lead to a large cut through the ECA and CCA when sampling the ICA centerline. (b) The bifurcation axis is defined as the line between the centerline bifurcation and the mass center of the ECA point and the ICA point. (c) Cross-section sampling perpendicular to the bifurcation axis in the bifurcation area and perpendicular to the centerline in the ICA, CCA, and ECA. (d) Bifurcation cross-section with overlay of automatic ECA and ICA contours (e) contours after combining ICA and ECA segmentation.}
    \label{fig:cross-section-placement}
    \label{fig:bifurcation-contours}
\end{figure}

For pseudo-label creation, we evaluated [0.3,~0.6,~1.2]~mm spacing between cross-sections. We also evaluated different cross-section placements in the bifurcation area. The first strategy samples all cross-sections perpendicular to the centerline. This leads to problems if the cross-sections cut through the bifurcation area (Fig.~\ref{fig:cross-section-placement}a). To prevent this, we defined the bifurcation axis as shown in Fig.~\ref{fig:cross-section-placement}b and sampled the bifurcation region along this axis (Fig.~\ref{fig:cross-section-placement}c).

The cross-sections along the bifurcation axis pose a challenge for the adversarial 2D U-Net as they are not centered on the vessel's centerline, leading to failed segmentation when the cross-section is centered on the bifurcation axis. To solve this, we used the adversarial 2D network to segment a cross-section centered on the ECA and ICA (Fig.~\ref{fig:bifurcation-contours}d). These contours are still incorrect, as the outer contour of the ICA touches the inner contour of the ECA and vice versa. To solve this, we joined the wall area prior to contour creation, which leads to the bifurcation contour shown in Fig.~\ref{fig:bifurcation-contours}e.

\paragraph{Training of the 3D Segmentation}
We used the nnU-Net framework with the 3D full resolution configuration and the default trainer. \cite{isensee2021nnunet}  For the network input, the 3D BB-MRI was resampled to an isotropic voxel size of 0.3~mm. The network was trained with the pseudo-labels described above.
We trained the nnU-Net with 5-fold cross-validation and ensured that all scans of a subject were in the same fold. For inference, we used the ensemble of all five folds.

\paragraph{Evaluation}
\begin{figure}
    \centering
    \includegraphics[width=1\linewidth]{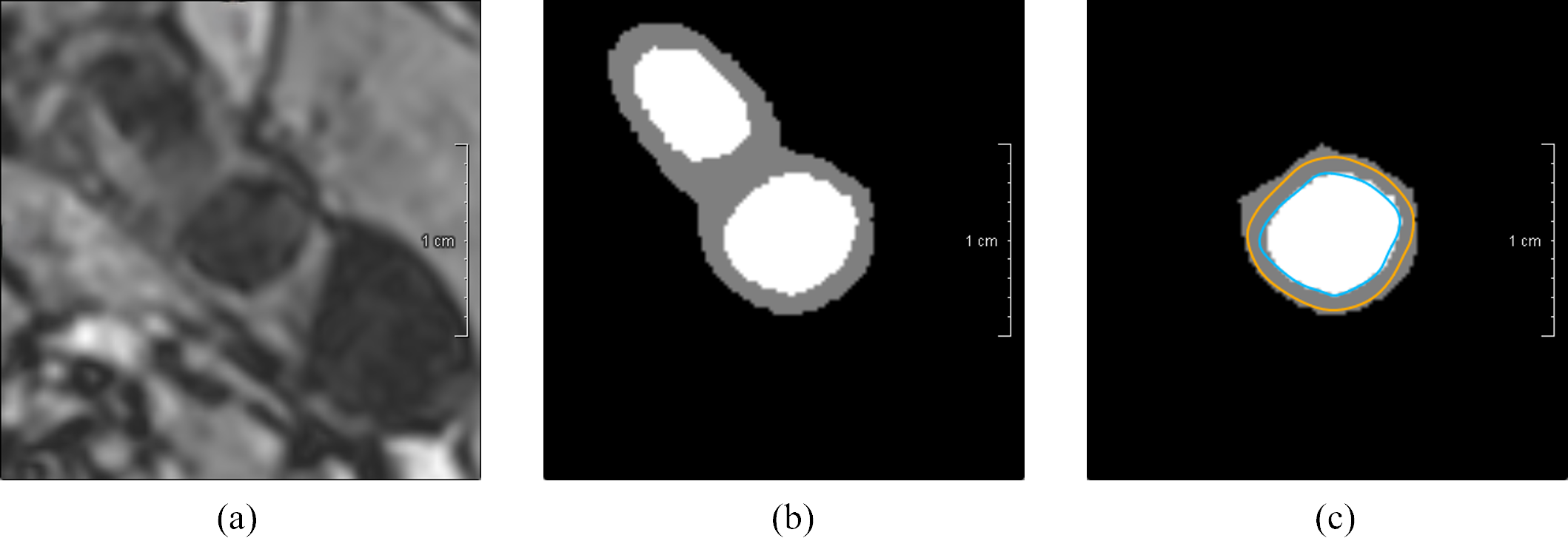}
    \caption{(a) Cross-section of a T1-weighted MRI with ICA and ECA visible (b) Cross-section of the automatic 3D segmentation with ECA and ICA segmentation (c) Automatic ICA segmentation and expert ICA contour annotation.}
    \label{fig:evaluation-bifurcation}
\end{figure}

We evaluated the 3D segmentation on the sparse 2D expert annotation contours. This requires post-processing if the ICA and ECA can be seen in the same cross-section (Fig.~\ref{fig:evaluation-bifurcation}a). If two vessels are present, the one that is further away from the center is labeled as background for evaluation. If the cross-section cuts through the bifurcation and shows a connected wall with two lumens (Fig.~\ref{fig:evaluation-bifurcation}b) this is not possible. To handle these cases, wall pixels are labeled as background if they are closer to the lumen that is not in the center. This leads to the corrected cross-section shown in Fig.~\ref{fig:evaluation-bifurcation}c which was used for evaluation.

As evaluation metrics, we used the mean of the symmetric average contour distance (ACD), symmetric Hausdorff distance (HD) and Dice-Sørensen coefficient (DSC) in the 2D slices. We also evaluated the performance of the different annotation planes shown in Fig.~\ref{fig:sparse-annotations}a. Cross-sections where the 3D segmentation did not segment a carotid artery were considered failed. We counted their occurrences and did not consider them for the calculation of the mean and median.

\paragraph{Model selection} 
To select the best strategy for pseudo-label creation, we performed a 5-fold cross-validation on the training set. We used the minimum number of failed cross-sections and the minimum mean HD as the decision criteria.

\section{Results}

\paragraph{Model Selection} Table~\ref{tab:cross-validation} shows that the sampling along the bifurcation axis for pseudo-label creation improves the segmentation metrics. Using a cross-section sampling distance (SD) of 1.2~mm results in lower DSC, and higher ACD and HD than a SD of 0.6~mm or 0.3~mm. Based on the decision criterion HD and the number of failed slices, we chose a sampling distance of 0.6~mm with the use of the bifurcation axis.

\begin{table}[h]
\caption{Ablation study on the influence of sampling strategy (bifurcation axis BA or centerline) and sampling distance (SD) for pseudo label creation evaluated with 5-fold cross-validation on the training set.}
\label{tab:cross-validation}
\begin{center}
\begin{tabular}{rr@{\quad}rrr@{\quad}rrr@{\quad}r}
\hline
&  & \multicolumn{3}{c}{Lumen}&  \multicolumn{3}{c}{Wall}  & Failed Slices/       \\
SD& BA  & $\mu$(ACD)    & $\mu$(HD) & $\mu$(DSC) & $\mu$(ACD)    & $\mu$(HD) & $\mu$(DSC) & Num Slices  \\ 
\hline
0.3&yes&\textbf{0.131}&\textbf{0.586}&\textbf{0.937}&\textbf{0.153}&0.851&\textbf{0.836}&26/2654\\
0.3&no&0.145&0.722&0.929&0.165&0.952&0.831&86/2654\\
0.6&yes&0.133&\textbf{0.586}&0.936&\textbf{0.153}&\textbf{0.845}&0.835&\textbf{22/2654}\\
0.6&no&0.156&0.751&0.930&0.176&0.985&0.830&42/2654\\
1.2&yes&0.142&0.616&0.935&0.165&0.887&0.832&25/2654\\
1.2&no&0.162&0.694&0.932&0.180&0.949&0.831&27/2654\\
\hline
\end{tabular}
\end{center}
\end{table}
\vspace{-2em}

\begin{figure}
    \centering
    \includegraphics[width=0.65\linewidth]{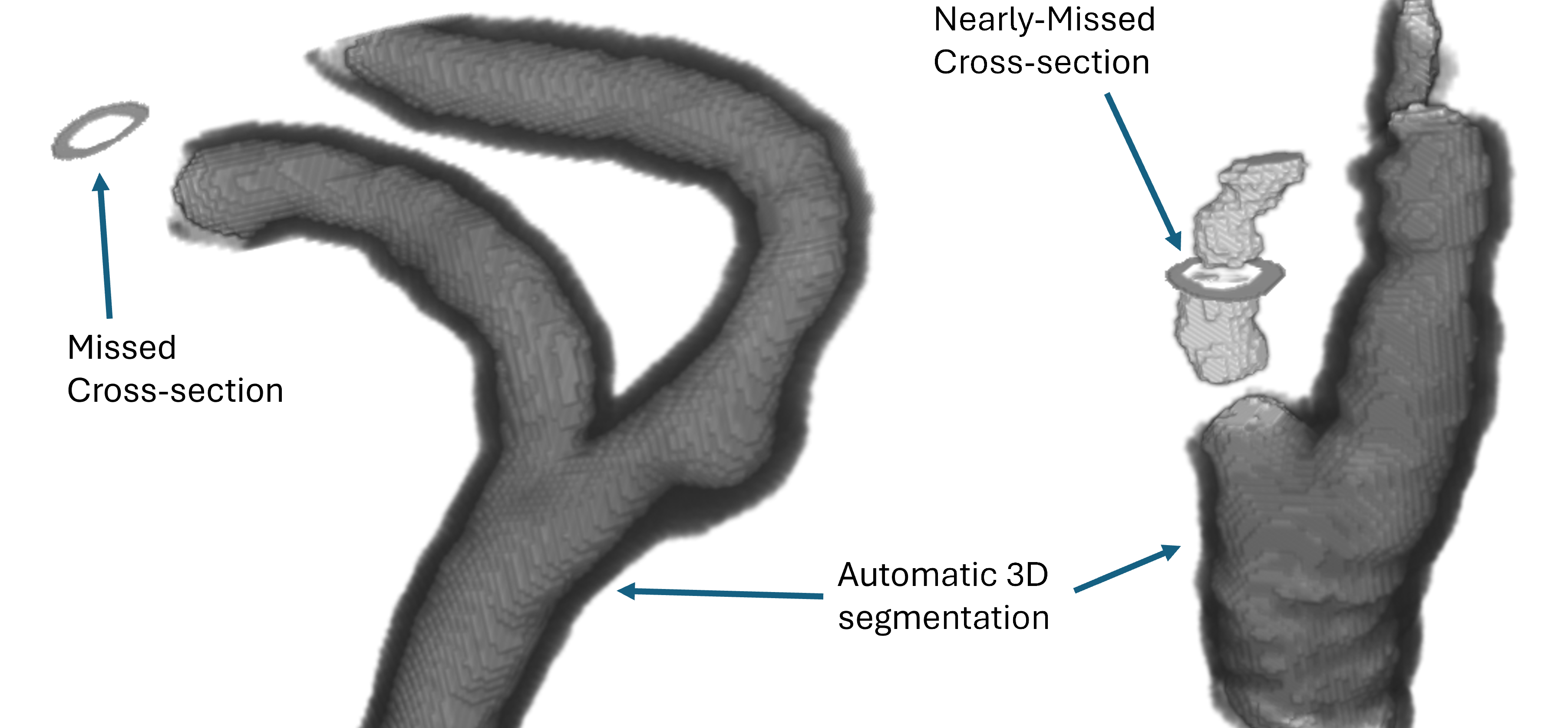}
    \caption{Incorrect segmentations in the ECA (Plane 8). On the left, the ECA segmentation does not continue up to the cross-section position. On the right, only parts of the ECA are segmented.}
    \label{fig:missed-cross-section}
\end{figure}

\paragraph{Segmentation results on the test set}
Table~\ref{tab:test_set_by_plane} shows the evaluation metrics for the automatic 3D segmentation evaluated on the annotated 2D cross-sections. The 3D nnU-Net segments the carotid artery in all but one annotated cross-section. The missed cross-section is positioned at plane 8 (in the ECA) (Fig.~\ref{fig:missed-cross-section}). 

\begin{table}[]
\caption{Evaluation of the 3D segmentation on the expert annotations of the test set.}
\label{tab:test_set_by_plane}
\begin{center}
\begin{tabular}{l@{\quad}rrr@{\quad}rrr@{\quad}r}
\hline
&  \multicolumn{3}{c}{Lumen}&  \multicolumn{3}{c}{Wall}  &Failed Slices/     \\
Plane & $\mu$(ACD)    & $\mu$(HD) & $\mu$(DSC) & $\mu$(ACD)    & $\mu$(HD) & $\mu$(DSC) & Num Slices  \\ 
\hline
Plane 1&0.082&0.416&0.961&0.109&0.619&0.869&0/37\\
Plane 2&0.107&0.558&0.955&0.144&0.853&0.845&0/36\\
Plane 3&0.422&1.402&0.869&0.462&2.234&0.772&0/36\\
Plane 4&0.129&0.530&0.935&0.147&0.828&0.839&0/37\\
Plane 5&0.126&0.548&0.932&0.135&0.739&0.838&0/39\\
Plane 6&0.102&0.452&0.936&0.119&0.596&0.839&0/36\\
Plane 7&0.145&0.553&0.902&0.130&0.657&0.804&0/33\\
Plane 8&0.149&0.486&0.892&0.182&0.675&0.772&1/35\\\hline
All planes&0.157&0.618&0.923&0.178&0.901&0.823&1/289\\
\hline
\end{tabular}
\end{center}
\end{table}

The ECA is segmented, but stops before the manual annotation. A second finding of the plane-wise evaluation is that the segmentation of cross-sections at plane 3 has the highest ACD and HD and the lowest DSC for lumen and wall segmentation. 

\begin{figure}
    \centering
    \includegraphics[width=0.95\linewidth]{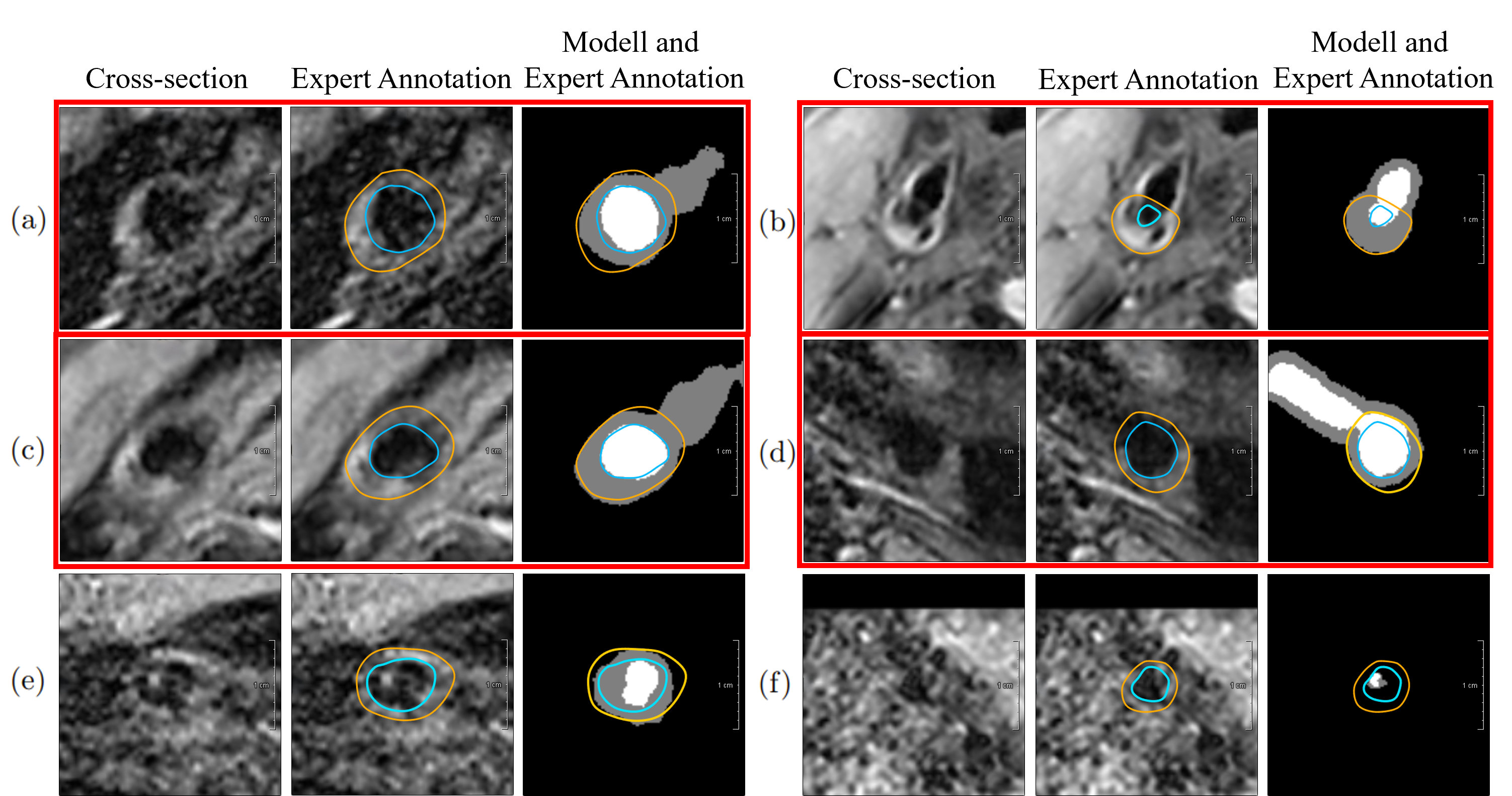}
    \caption{Cross-sections with the lowest DSC, highest ACD, and highest HD. Cross-sections marked in red are correctly segmented, but the evaluation on the sparse annotations results in poor evaluation metrics.}
    \label{fig:outlaiers}
\end{figure}

Looking at the example cross-sections with the worst evaluation metrics shown in Fig.~\ref{fig:outlaiers}, one can see that cross-sections (a)-(d) are at plane~3, and the bad performance metrics are caused by the ECA segmentation being present in the 2D evaluation cross-section. Cross-section (e) shows a strong flow artifact that was labeled as wall by the automatic segmentation. Cross-section (f) shows plane 8 (in the ECA). The automatic 3D segmentation can be seen on the right of Fig.~\ref{fig:missed-cross-section}, the network did not segment the ECA properly.

\begin{figure}
    \centering
    \includegraphics[width=0.8\linewidth]{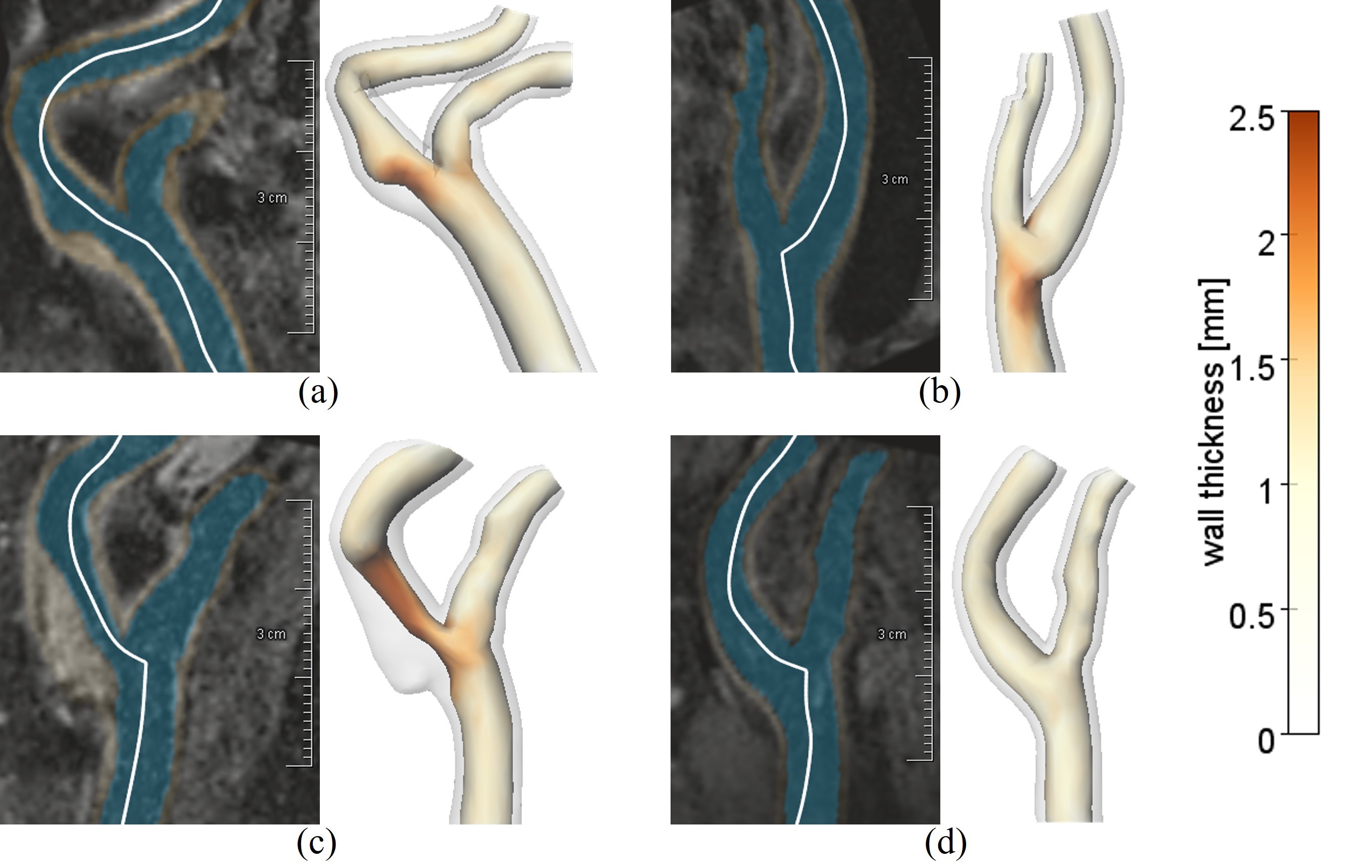}
    \caption{Example segmentations produced by the network. On the left, a curved multiplanar reconstruction with the lumen segmentation (blue) and the wall segmentation (orange) are given. On the right, the corresponding color-coded visualization of the vessel wall thickness is demonstrated. (a) Artery with high tortuosity; (b) Artery with a mild stenosis; (c) Artery with a moderate to severe stenosis; (d) Artery of a healthy subject.}
    \label{fig:wallthickness}
\end{figure}
Fig.~\ref{fig:wallthickness} shows successfully segmented carotid arteries and a visualization of the wall thickness. The examples show a different degree of stenosis and different carotid artery shapes.

Table~\ref{tab:generalization} shows the evaluation metrics for different datasets. The model failed to localize lumen and wall in one case for healthy subjects. This happened at plane 7, the most distal evaluation cross-section in the ICA. In the carotid artery segmentation challenge dataset, it failed to localize lumen and wall in 8 of 4189 cases. The model achieves lower ACD and HD and higher DSC for healthy subjects. The model does not perform as well on the stenosis test set. This is mainly caused by one stenosis cross-section, in which the automatic  evaluation where the ECA is present. The model is able to segment the test set of the 2021 Carotid Artery Vessel Wall Segmentation Challenge \cite{Challenge2021} and achieves a lower mean HD than on the test set.

\begin{table}
\caption{Evaluation of the networks performance on different test sets.}
\label{tab:generalization}
\begin{center}
\begin{tabular}{l@{\quad}rrr@{\quad}rrr@{\quad}r}
\hline
&  \multicolumn{3}{c}{Lumen}&  \multicolumn{3}{c}{Wall} & Failed Slices/       \\
Dataset & $\mu$(ACD)    & $\mu$(HD) & $\mu$(DSC) & $\mu$(ACD)    & $\mu$(HD) & $\mu$(DSC) & Num Slices  \\ 
\hline
Test set                            &0.157&0.618&0.923&0.178&0.901&0.823&1/289\\
Healthy                             &0.139&0.619&0.940&0.171&0.818&0.788&1/155\\
Stenosis                            &0.411&1.254&0.803&0.359&1.536&0.808&0/10\\
Challenge \cite{Challenge2021}      &0.161&0.595&0.908&0.177&0.767&0.761&8/4189\\\hline
\end{tabular}
\end{center}
\end{table}

\section{Discussion}
We presented an iterative approach to train a nnU-Net for the 3D segmentation of the carotid artery wall on sparse cross-sectional annotations. We create 3D pseudo-labels, using an adversarial 2D network and the centerline. In contrast to our prior work \cite{rahlfs2024carotid}, which evaluated the 2D segmentation of the carotid artery wall, the proposed 3D network does not require the vessel centerline as input. 
This approach was tested and assessed with annotations created for clinical studies.
The performance of the segmentation depends on the quality of the pseudo-labels, and our experiments suggest that special handling of the bifurcation can improve the final segmentation performance. The distance, with which the centerline is sampled for pseudo-label creation, has a smaller effect than the sampling direction relative to the bifurcation. For the evaluation dataset, a sampling distance of 0.6~mm creates the best results, which corresponds to the slice thickness suggested for carotid wall imaging by the AJNR \cite{saba2018carotid}. We assume that the performance of the 2D adversarial network has a large effect on the segmentation performance, and further advances in the 2D segmentation might improve the 3D segmentation.
In some cases, the distal areas of the ECA and ICA are not properly segmented, as seen in Fig.~\ref{fig:missed-cross-section}. We do not consider this to be a problem in clinical practice, as the areas of increased vessel wall thickness are most prominent in the carotid bulb and the bifurcation area.

We used sparse annotations to evaluate the 3D segmentation and achieved good results. However, the proposed evaluation poses the problem that only the CCA, ICA, or ECA are segmented in one 2D cross-section, even if another vessel is visible in the cross-section (Fig.~\ref{fig:sparse-annotations}b). This shows most prominently in the bifurcation area, which leads to high ACD and HD for the evaluation of Plane 3, as seen in Table~\ref{tab:test_set_by_plane}. Fig.~\ref{fig:outlaiers}a-d also show that these outliers are not caused by incorrect 3D segmentation but by the 2D annotations that are used for evaluation. So it might be necessary to provide expert-corrected 3D annotations of the test set for a better evaluation of our method.

The 3D wall segmentation can be used to extract advanced biomarkers such as 3D plaque properties, distributions of wall thickness, or radiomics features. 

\paragraph{Limitations}
We used sparse annotations for the evaluation of the method, which does not allow a proper evaluation of the segmentation performance in the bifurcation area and in the ECA, which is underrepresented in the sparse annotations. 

The 3D segmentation of the carotid artery wall does not allow for a direct calculation of the plaque volume as the plaque is not segmented.

We did not perform an evaluation of the generalization to different arteries, field strengths, or higher degrees of stenosis. 
%
%

\end{document}